\pdfoutput=1

\documentclass[11pt]{article}

\usepackage{acl}
\usepackage{booktabs} 
\usepackage{xcolor,colortbl}
\usepackage{parcolumns}
\usepackage{adjustbox}
\usepackage{float}

\usepackage{times}
\usepackage{latexsym}
\PassOptionsToPackage{export}{adjustbox}
\usepackage[export]{adjustbox}

\usepackage[T1]{fontenc}
\usepackage{algorithm}
\usepackage{algpseudocode}
\usepackage{listings}
\usepackage{xcolor}

\usepackage[utf8]{inputenc}
\usepackage{graphicx}
\usepackage{tabularx}
\usepackage{microtype}
\usepackage{makecell}
\usepackage{amssymb}
\usepackage{hyperref}
\usepackage{inconsolata}
\usepackage{tablefootnote}
\usepackage{makecell}

\NewDocumentCommand{\xudong}{ mO{} }{\textcolor{blue}{\textsuperscript{\textit{Xudong}}\textsf{\textbf{\small[#1]}}}}

\title{Unveiling Narrative Reasoning Limits of Large Language Models with Trope in Movie Synopses}

\author{
Hung-Ting Su$^{1}$\thanks{Equal contribution.} \quad Ya-Ching Hsu$^{1}$\footnotemark[1] \quad Xudong Lin$^{2}$ \quad Xiang-Qian Shi$^{1}$ \\ \textbf{Yulei Niu}$^{2}$ \quad \textbf{Han-Yuan Hsu}$^{1}$ \quad \textbf{Hung-yi Lee}$^{1}$ \quad \textbf{Winston H. Hsu}$^{1,3}$
\\$^{1}$National Taiwan University \qquad $^{2}$Columbia University \qquad $^{3}$Mobile Drive Technology
}

\begin{document}
\maketitle
\begin{abstract}
Large language models (LLMs) equipped with chain-of-thoughts (CoT) prompting have shown significant multi-step reasoning capabilities in factual content like mathematics, commonsense, and logic. However, their performance in narrative reasoning, which demands greater abstraction capabilities, remains unexplored. This study utilizes tropes in movie synopses to assess the narrative reasoning abilities of state-of-the-art LLMs and uncovers their low performance. We introduce a trope-wise querying approach to address these challenges and boost the F1 score by 11.8 points. Moreover, while prior studies suggest that CoT enhances multi-step reasoning, this study shows CoT can cause hallucinations in narrative content, reducing GPT-4's performance. We also introduce an Adversarial Injection method to embed trope-related text tokens into movie synopses without explicit tropes, revealing CoT's heightened sensitivity to such injections. Our comprehensive analysis provides insights for future research directions. Code available: \url{https://github.com/Shelley1214/Trope}

\end{abstract}

\section{Introduction}\label{sec:1}


\begin{figure}[ht]
    \centering
    \includegraphics[width=\linewidth]{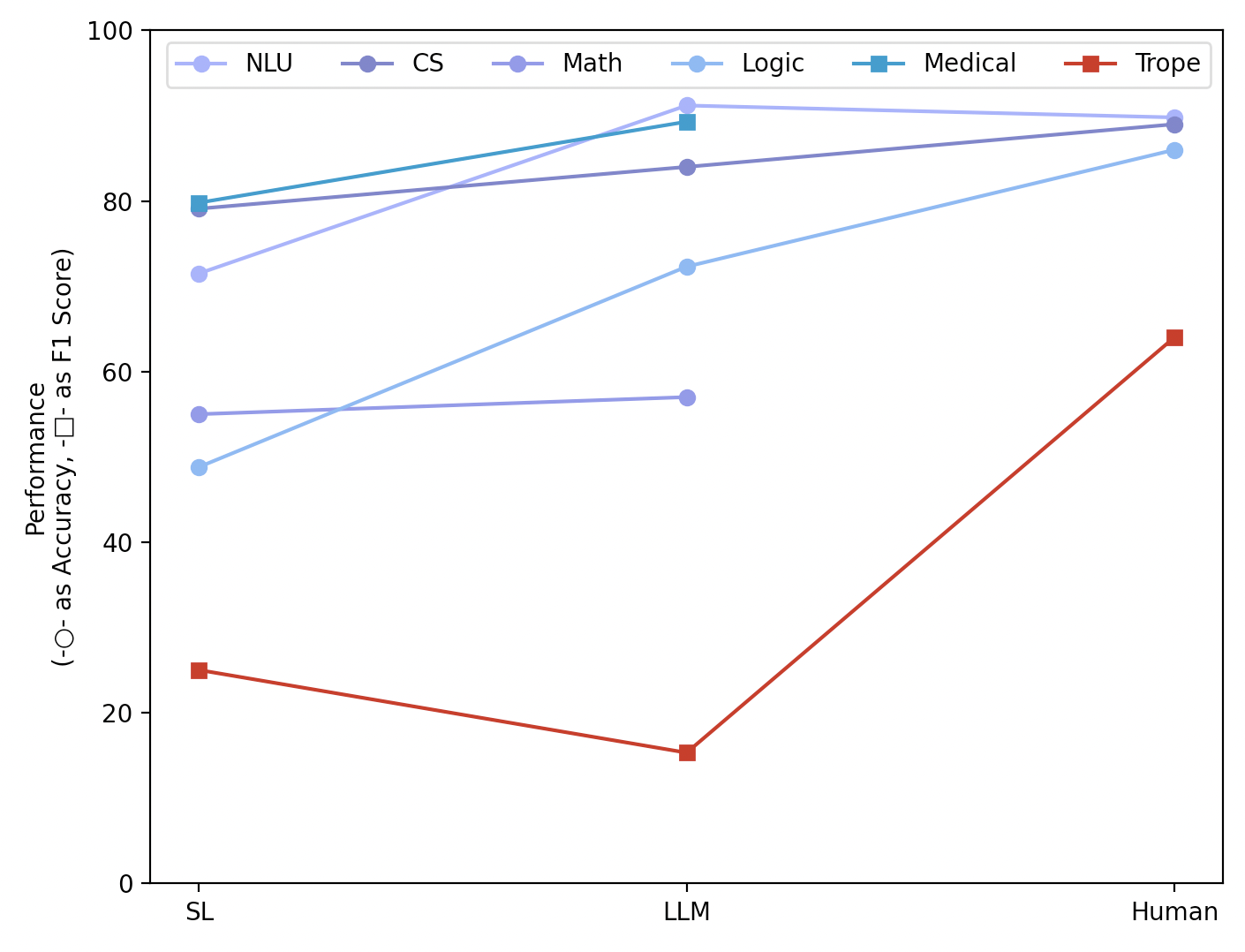}
    \caption{While LLMs have revolutionized NLP reasoning, surpassing previous supervised learning (SL) methods and even reaching human-level performance on some tasks, their limitations become apparent when tested against the Trope dataset.
    NLU: Natural Language Understanding, CS: Commonsense. Check Section \ref{sec:1} and \ref{subsec:2:llmreason} for details.}
    \label{fig:figure1}
\end{figure}





Large Language Models (LLMs) \cite{gpt3,gpt4,touvron2023llama} have shown their significant reasoning skills in few-shot manners with billion-scale parameters. Chain-of-thoughts (CoT) \cite{wei2022chaincot} extends LLMs' reasoning capabilities by introducing stepwise reasoning. This approach resembles ``slow thinking'' in cognitive science \cite{kahneman2011thinkingfastandslow}, enabling LLMs to break down complex tasks into a sequence of simpler tasks, facilitating progressive processing and integration of results. It enhances performance across various tasks, including arithmetic calculations \cite{cobbe2021traininggsm8k} and domain-specific reasoning \cite{liu2023exploring}. 

\begin{figure*}[ht]
  \includegraphics[width=\textwidth]{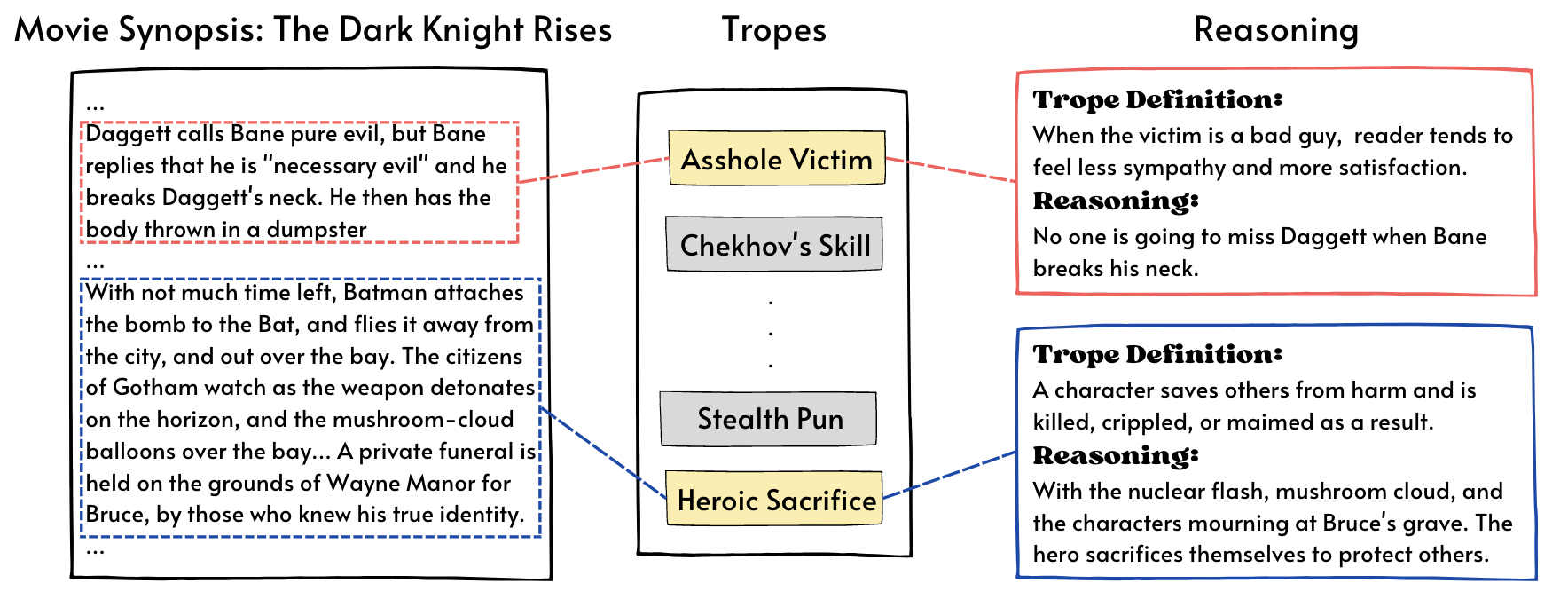}
\caption{
%
Trope in Movie Synopses (TiMoS) requires the abstraction of narrative reasoning beyond physical presentation. For example, themes of justice (red block) or sacrifice (blue block) extend beyond death. TiMoS also explores connections between seemingly unrelated ideas, such as Batman's departure and his efforts to save the city (blue block).
}


 \label{fig:overview}
\end{figure*}

Unlike existing LLM analyses, which primarily focus on factual reasoning and have demonstrated significant advances in LLMs and CoT processes for factual reasoning, this work presents the first investigation of LLMs using tropes in movie synopses. It reveals that LLMs and CoT lack \textit{narrative reasoning} reasoning skills, as shown in Figure \ref{fig:figure1}. 
Unlike factual reasoning, which is based on logical deductions and objective data \cite{zhang2020answerfact}, narrative reasoning presents distinct challenges by requiring a deep understanding of event sequences and extensive world knowledge \cite{narrativejiayang2024eventground}. Since narratives encapsulate human behavior, beliefs, and motivations beyond displayed contexts \cite{piper2021narrative}, the trope in movie synopses task introduces several novel challenges and perspectives beyond previous work, as shown in Figure \ref{fig:overview}. 

First, trope understanding requires LLMs to understand concepts that are not physically present or directly observable. For example, ``Heroic Sacrifice'' involves a character choosing to give up their own life or well-being for the greater good or to protect others. The concept goes beyond the physical act of sacrifice to encompass the thematic implications of altruism, selflessness, and the value of individual life versus collective well-being. This trope is often portrayed as the death of a character, but the death of the character does not entail the trope (it could even be the exact opposite, as shown in Figure \ref{fig:overview}). Existing benchmarks such as math reasoning \cite{cobbe2021traininggsm8k} or natural language inference (NLI) \cite{wang2019superglue} provide concrete and observable context. For math reasoning, concepts and operations are transparent. While NLI involves abstract thinking to some degree to infer relationships between sentences and understand implied meanings, the scope is confined to linguistic and logical reasoning within specific textual contexts. NLI tasks do not demand the same level of thematic interpretation, symbolic analysis, or connection of ideas across disparate narrative elements that trope understanding does. 

Second, comprehending tropes requires LLMs to make connections between seemingly unrelated ideas. ``Heroic Sacrifice'' requires viewers or readers to connect the act of sacrifice with broader themes or messages of the narrative, such as freedom, love, or redemption. These themes and messages not only are hard to be observed but also appear to be completely unrelated. For example, a character might subtly express their love for freedom in a casual conversation, which later informs their ultimate sacrifice. However, randomly grasping several ``seemingly unrelated ideas'' without carefully reasoning between the ideas could result in hallucination. For example, ``a character A loves B'' and ``the character A dies in an accident in front of B'' could be unrelated to Heroic Sacrifice despite seeming to have the elements of death and love. Meanwhile, existing benchmarks do not require such a capability. For example, while logic reasoning \cite{liu2023logiqa} and commonsense reasoning \cite{CommonsenseQA} assess AI's logical reasoning and commonsense knowledge, they do not require the nuanced thematic interpretation and narrative analysis needed to understand tropes like Heroic Sacrifice. ``Heroic Sacrifice'' involves integrating complex narrative themes and character motivations, a level of abstract thinking and interpretation beyond the structured challenges of logical reasoning and the common knowledge queries of commonsense reasoning.

Exploring narrative reasoning capabilities is beneficial for both LLM research and application development. It challenges modern LLMs by requiring abstraction from narrative contexts, a realm beyond previous research. This exploration sheds light on LLM behaviors under specific circumstances and encourages the development of more reliable applications to mitigate hallucinations. 
In light of this, we first investigate LLMs' reasoning capability by revisiting an existing Trope in Movie Synopses (TiMoS) dataset \cite{chang2021situationtimos} and discover that it remains a very challenging task even for current LLMs with a well-engineered prompting pipeline. As Figure \ref{fig:figure1} shows, advanced LLMs such as GPT-4, ChatGPT, and fine-tuned LLaMa-2 perform poorly, only achieving F1 scores at the level of random guessing even when they are equipped with CoT \cite{wei2022chaincot} prompts. This suggests that state-of-the-art LLMs, despite dominating various benchmarks, do not carry reasoning skills for challenges in trope understanding tasks. 

Furthermore, we address the challenge by reframing the TiMoS task as trope-wise querying, where each LLM query inputs a single trope, significantly enhancing LLM performance. As a result, performance improves by 11.8 points on the F1 score, surpassing the supervised state-of-the-art results proposed by TiMoS \cite{chang2021situationtimos}. This strategy opens a new pathway for tackling complex reasoning tasks by decomposing multiple concepts into a single concept within an LLM query. 

In addition to assessing trope reasoning capabilities, we reveal CoT's tendency for hallucination and show that it does not always enhance reasoning compared to vanilla prompting. Furthermore, CoT can increase LLMs' susceptibility to adversarial inputs. 
Specifically, inspired by prior reading comprehension studies \cite{jia2017adversarialsquad}, we devise an Adversarial Injection that inserts trope-related text tokens without explicit trope introduction into a movie synopsis, aimed at gauging whether LLMs are misled. 
The propensity for hallucination tendency is underscored by: 
(1) a stark decline in LLM precision when employing CoT, 
(2) Adversarial Injection significantly misleads LLMs through keyword and pattern recognition, especially when CoT is equipped, and 
CoT generates accurate responses with erroneous rationales. We also provide a comprehensive analysis, highlight the challenges associated with TiMoS, and offer insights for future LLM research and applications.

The contributions of this paper are as follows:
\begin{itemize}
    \item This work presents the novel effort to utilize tropes in movie synopses to scrutinize LLMs' trope reasoning skills and offer new perspectives compared to previous analyses.
    \item Our investigation reveals that advanced LLMs, including ChatGPT, GPT-4, and LLaMa-2, even with CoT or fine-tuning, lack the skills necessary for understanding tropes (Section \ref{subsec:4:2:mainresults}).
    \item We significantly improve LLMs' performance on the TiMoS task by 11.8 points in F1 score through trope-wise querying. (Section \ref{subsec:4:3:binaryresults})
    \item We expose the limitations of CoT prompting, revealing its propensity for hallucination and increased vulnerability to adversarial inputs. (Section \ref{subsec:4:cotresults})
\end{itemize}


\section{Related Work}\label{sec:2}

\subsection{Large Language Models (LLMs)}
Large Language Models (LLMs) \cite{gpt3,gpt4,touvron2023llama} have demonstrated their dominant power in various NLP tasks. Compared to traditional \textit{``smaller''} language models \cite{BERT,gpt2-radford2019language,T5_JMLR:v21:20-074} that require fine-tuning to adapt to downstream tasks, LLMs carry much more parameters (100 billion scale) and do not require task-specific fine-tuning to tackle downstream tasks. Without the requirement of massive human annotation, LLMs have reached supervised state-of-the-art or even human-level performance. When equipped with in-context learning, where LLMs implicitly learn from few examples without updating parameters, and advanced prompting techniques such as CoT \cite{wei2022chaincot}, LLMs even dominate tasks that were considered challenging and require new paradigms to conquer. 

\subsection{LLM Reasoning}\label{subsec:2:llmreason}
Natural Language Understanding (NLU) \cite{wang2018glue,wang2019superglue} is a collection of tasks that examine machines' capability of understanding language in general, flexible, and robust manners. 
Commonsense reasoning \cite{talmor2019commonsenseqa,talmorcommonsenseqa2,ismayilzada2023crow} requires not only understanding the context but also referring to commonsense knowledge. Both NLU and commonsense reasoning are challenging for task-specific supervised models, even with pre-trained knowledge. A recent research \cite{singh2023mindscores1} reported that GPT-4 \cite{gpt4} surpassed human performance on SuperGLUE \cite{wang2019superglue} NLI benchmark and shortened the gap between human and machine to 5.0 accuracy on CommonsenseQA \cite{talmor2019commonsenseqa} commonsense reasoning dataset. Besides NLI and commonsense, some recent research also indicates that LLMs can serve as domain-specific experts. \cite{liu2023exploring} demonstrated GPT-4+CoT's reasoning skills that surpass previous supervised state-of-the-art by 10 F1 score in medical domain reasoning \cite{miura-etal-2021-improving}. \cite{wei2022chaincot} showed that CoT significantly boosted GPT-3's \cite{gpt3} performance and outperformed supervised state-of-the-art by 2.0 points on math dataset \cite{cobbe2021traininggsm8k}. 
\cite{liu2023evaluating} examined ChatGPT and GPT-4's skills in logical reasoning and reported that GPT-4 significantly outperformed supervised state-of-the-art by 23.5 of accuracy. 

\subsection{Tropes}
In recent years, tropes have received attention from the research community for developing multimedia content creation tools \cite{smith2017harnessing,chou2023talestream} or serve as a testbed of machine reasoning skills \cite{chang2021situationtimos,su2021truman}. Previous work \cite{chang2021situationtimos,su2021truman} confirmed that understanding tropes in movies requires deeper cognition skills compared to existing ones and tested supervised models such as BERT \cite{BERT} or graph neural network-based multi-step reasoning model \cite{DBLP:conf/nips/PalmPW18}. A significant gap was observed between the state-of-the-art supervised model (25 F1) and the human (64 F1) on Tropes in Movie Synopses (TiMoS) \cite{chang2021situationtimos} dataset. This study employs it as a testbed for modern LLMs, which have excelled in multiple challenging NLP tasks across various domains. We investigate the narrative reasoning capabilities of LLMs and CoT, uncovering their tendency for hallucination.



\section{Narrative Reasoning with TiMoS}\label{sec:3}

\subsection{Experimental Setup}\label{subsec:3:setup}


\paragraph{Task.} TiMoS \cite{chang2021situationtimos} is a multi-label classification task where an LLM inputs a movie synopsis and predicts if specific tropes exist. An LLM is given an article and a set of tropes, and the model needs to sift through the content and select the relevant tropes. This demands a comprehensive analysis of the entire plot, necessitating a holistic understanding of the interwoven elements within the text.

\paragraph{Trope-wise Querying.} While humans can process multiple concepts from a narrative simultaneously \cite{Huang2011}, LLMs might lack the same level of skill. In this approach, an LLM is given an article and a single trope, performing $N$-label classification with $N$ Trope-wise Querying. This allows the LLM to focus on a single trope and pay attention to its elements.

\paragraph{Prompting.} Two prompting approaches are considered\footnote{Please check Appendix for more details.}. Firstly, the Base Prompting utilizes standard zero-shot prompting, where a model is given an article and asked to assess its relevance to a particular trope. Secondly, the CoT \cite{wei2022chaincot} Prompting approach. 
For the input, we segmented the synopsis into sentences, each annotated with a numerical identifier to aid the model's reasoning. For the output, the model is asked to identify selected tropes or provide a binary answer ("yes" or "no" in the Trope-wise Querying setting). To encourage stepwise trope reasoning, we request additional outputs from the LLMs: (1) Trope Definition: a brief explanation of the trope, (2) Reasoning: why the trope is depicted in the plot, and (3) Relevant Paragraphs: numerical identifiers of sentences where evidence is found. The model is provided with an illustrative example to guide its analysis of tropes within articles.


\paragraph{Fine-Tuning.}
 To verify whether the model's challenge lies in understanding tropes or stems from the absence of supervised learning on this corpus, we additionally fine-tuned an LLaMa-2 \cite{touvron2023llama} LLM with instructions for both querying way, which represents an approximate upper limit of current LLMs' capabilities in training-free settings. In the multi-label setting, the input includes the plot and all relevant tropes, with the output being the identified tropes. In trope-wise querying, the input consists of a plot and a single trope, with the output indicating whether the trope is present or not. While we acknowledge the performance gap between the 7B model and GPT-4, this is the best available option for our analysis.\footnote{Please check Appendix \ref{apdx:finetune_detailed} for more fine-tuned details.}

\paragraph{Large Language Models.}
In our thorough examination of LLMs through prompting, we centered our attention on the most cutting-edge models available: GPT-4 \cite{gpt4}, ChatGPT, and LLaMa-2-7B-chat \cite{touvron2023llama}. For GPT-4 and ChatGPT, we use the APIs (gpt-3.5-turbo, gpt-3.5-turbo-16k, and gpt-4 models) provided by OpenAI. To ensure precision, we set the temperature to 0, guaranteeing deterministic outputs from the models, consistently yielding the word with the highest probability. We utilized the checkpoint provided by Meta for LLaMa-2 experiments and used default settings for other hyperparameters. In the fine-tuned setting, we fine-tune LLaMa-2-7B for multi-label querying and trope-wise querying.

\subsection{LLMs Struggle Reasoning TiMoS}\label{subsec:4:2:mainresults}
The first block of Table \ref{table:2} demonstrates LLMs' trope understanding results (fifth to last row) on TiMoS \cite{chang2021situationtimos} compared to traditional supervised baselines (first to fourth row).\footnote{Please check Appendix \ref{apdx:baseline} for more baseline details.} Supervised state-of-the-art \cite{chang2021situationtimos} can reach 25.00 of F1 score and is still far behind human performance (64.87 F1). 
The fifth and sixth rows display our exploration of TiMoS tasks using the CoT approach — where we tasked the model with predicting all relevant tropes, reasons, and related paragraphs at once, just like how human and supervised models do. 
Surprisingly, we observed significantly low results with LLMs. Even dominating GPT-4+CoT still obtained an 15.33 F1 score. This highlights the inherent narrative reasoning challenges LLMs encounter in trope understanding, necessitating the decomposition of plot elements and stepwise reasoning to integrate results. In addition, ChatGPT faces greater difficulty compared to its successor, GPT-4, despite being considered extremely powerful and already tackled various factual reasoning benchmarks. 
In addition, while expressing ideas clearly without adjusting prompts, ChatGPT tended to generate responses outside the choices. This highlights the necessity for meticulous prompt engineering to ensure comprehension for ChatGPT, especially when GPT-4 already exhibits sufficient understanding. Moreover, while the performance of LLaMa-2 fine-tuning remains unsatisfactory, it surpassed that of GPT, indicating potential scarcity in multi-label task corpora for LLMs.

\begin{table}[ht]
\centering
\begin{adjustbox}{width=\linewidth}
\begin{tabular}{lllccc}
\toprule[1pt]
  \makecell{Trope-wise \\ Querying} & CoT & \textbf{Models} & \textbf{F1}&  \textbf{P} \tablefootnote{Baselines reported by TiMoS \cite{chang2021situationtimos} (first block) employs mean average precision(mAP) across various thresholds, while our approach (other blocks) directly predicts the answer and opts for Precision.} & \textbf{R}  \\
\cmidrule(lr){1-6}
& \makecell{$\times$}& Random$^{*}$ & 13.97 & 8.14 & -\\
& \makecell{$\times$}&BERT$^{*}$ & 23.97 & 17.26 &- \\
& \makecell{$\times$}&MulCom$^{*}$ & 25.00 & 18.73 & -\\ 
& \makecell{$\times$} &Human$^{*}$ & 64.87 & 65.77 & 63.98 \\
\cmidrule(lr){2-6}
\makecell{$\times$}& &\multicolumn{4}{l}{\cellcolor[gray]{0.8}\textbf{Prompt Results}}  \\
& \makecell{\checkmark} &ChatGPT & 13.19 & 24.60 & 9.01 \\
& \makecell{\checkmark} &GPT-4 & 15.33  & 22.80 & 15.33\\ 
&&\multicolumn{4}{l}{\cellcolor[gray]{0.8}\textbf{Fine-Tuned Results}} \\
& \makecell{$\times$} & LLaMa-2 & 16.37 &  19.66 &  14.03 \\
\cmidrule(lr){1-6}
&&\multicolumn{4}{l}{\cellcolor[gray]{0.8}\textbf{Prompt Results}} \\
& \makecell{$\times$} &LLaMa-2 &  15.54 &  8.44 & \textbf{97.72}\\
\makecell{\checkmark} & \makecell{$\times$} &ChatGPT  &  18.89 &   17.96 &   19.93 \\ 
& \makecell{$\times$} &GPT-4& \textbf{27.1} & \textbf{ 19.5} &  44.38 \\
& \makecell{\checkmark} &ChatGPT  &  20.25 & 11.8 &  71.24 \\
\bottomrule [1pt]
\end{tabular}
\end{adjustbox}
\caption{
The first block (w/o Trope-wise Querying) shows that all SOTA LLMs lack reasoning skills tackling TiMoS (Section \ref{subsec:4:2:mainresults}). The second block (w/ Trope-wise Querying) shows remarkable improvement with Trope-wise Querying equipped (Section \ref{subsec:4:3:binaryresults}). Check Section \ref{subsec:3:setup} for setups. $^{*}$: extracted from \cite{chang2021situationtimos}.
}
\label{table:2}
\end{table}

\subsection{Trope-wise Querying Improves LLMs}\label{subsec:4:3:binaryresults}

As demonstrated in the last block of Table \ref{table:2}, trope-wise querying, where we query the LLM for a single trope, significantly boosts LLM performance and enables GPT-4 to surpass the state-of-the-art supervised model. This is because it no longer needs to handle various concepts that require focusing on different parts of the plot and different comprehension paths. This finding indicates that LLMs do not carry the reasoning capability of processing multiple different concepts at once. The finding also echos concurrent paper \cite{multilabel} that emphasizes the crucial role of fine-tuning to address this limitation and enhance GPT's performance, specifically in the context of multi-label classification. It also reveals a direction to improve LLMs further and suggests a prompt engineering approach to the aspect of applications. Furthermore, Llama-2-7B and ChatGPT perform significantly worse than GPT-4 despite their success in other NLP tasks, indicating that GPT-4 has a deeper grasp of the complexities involved in the problem. Specifically, LLaMa-2 tends to predominantly predict ``yes,'' suggesting a potential limitation in comprehending higher-level dynamics for open-source LLMs available at the moment. 

\begin{table}
\centering
\begin{adjustbox}{width=\linewidth}
\begin{tabular}{llccc}
\toprule[1pt]

CoT & \textbf{Model} & \textbf{F1} & \textbf{P} & \textbf{R} \\
\hline
&\multicolumn{4}{l}{\cellcolor[gray]{0.8}\textbf{Prompt Results}} \\
  \makecell{$\times$} & ChatGPT& 20.57 & 22.57 & 18.9\\
 \makecell{$\times$} & GPT-4 & 29.59 & 20.86 &  50.87 \\ 
  \makecell{\checkmark} & ChatGPT & 24.99 & 15.09 & 72.67 \\
   \makecell{\checkmark} & GPT-4 & 27.91 &  19.19 &   51.16 \\  
 & \multicolumn{4}{l}{\cellcolor[gray]{0.8}\textbf{Fine-Tuned Results}} \\
  \makecell{$\times$} & LLaMa-2 & 19.93 & 15.52 &  27.84  \\
\bottomrule [1pt]
\end{tabular}
\end{adjustbox}

\caption{
Trope-wise Query analysis with different LLMs and prompting strategies (Section \ref{subsec:4:4:cot}) using a subset of 100 articles and 20 plots.
}
\label{table:3}
\end{table}

\subsection{Challenges of Chain-of-Thoughts (CoT)}\label{subsec:4:cotresults}

\subsubsection{CoT Diminishes GPT-4 Performance}\label{subsec:4:4:cot}
Many works of literature, such as Program CoT \cite{gao2022pal2}, Symbolic Reasoning \cite{suzgun2022challenging}, or Math reasoning \cite{cobbe2021traininggsm8k}, suggest that chain-of-thought prompting and its variants significantly improve reasoning tasks that require stepwise comprehension.
We conduct an experiment equipping ChatGPT and GPT-4 with CoT, and the results are shown in Table \ref{table:3}. Due to budget limitations, we sampled a subset with 100 synopses and 20 tropes for comparison. Different from previous work \cite{yao2023tree,TrueDetective}, we observe that chain-of-thought, while remarkably improving ChatGPT performance, does not work well on GPT-4 here. As chain-of-thoughts mainly boosts recall and slightly degrades precision as a trade-off, and ChatGPT has much lower recall than GPT-4 with base prompting, it might reveal that GPT-4 implicitly knows stepwise reasoning for tropes. 

\begin{figure}[h]
    \centering
    \includegraphics[width=\linewidth]{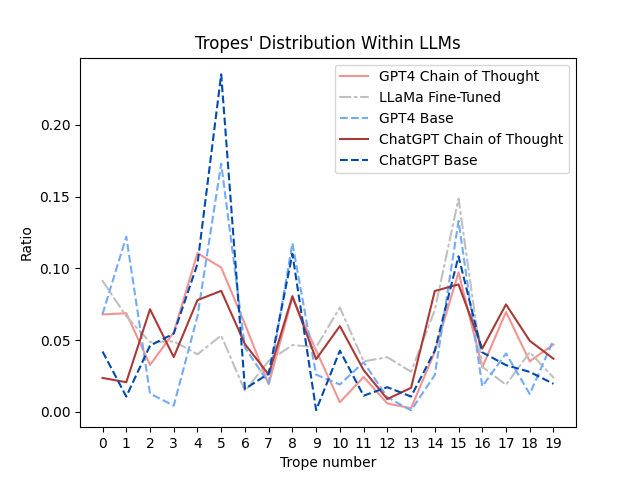}
    \caption{The distribution of each trope in forecasting a ``yes'' outcome varies across the five binary classification results within the subset. See Appendix for more results.}
    \label{fig:subset_distribution}
\end{figure}

Results in Table \ref{table:3} also suggests that there is room for improvement in narrative reasoning, even more than the widely used CoT approach. 
Additionally, as CoT boosts the recall at the price of precision, it could over-conceive the synopses and aggravate hallucination issues, which should be carefully inspected for truth-worthy applications. 
Furthermore, as illustrated in Figure \ref{fig:subset_distribution}, we note that the distributions of ChatGPT Base and GPT-4 Base closely align, as do those of ChatGPT CoT and GPT-4 CoT.
This indicates that Base and CoT represent two distinct modes of thinking. Regardless of the mode employed, neither effectively addresses the problem. The consistent tendency of both models to predict specific tropes suggests a potential blind spot in GPT models regarding tropes, implying that certain tropes may be consistently overlooked or inadequately emphasized.


\begin{table*}[h]
    \centering
    \begin{tabular}{ |c|cc|cc |} 
    \hline
    
     Trope & Component& Remove & Base  & CoT \\ 
     \hline
     Blatant Lies & obvious + lie & obvious & 63.33& 96.67 \\ 
     \hline
     Kick The Dog & does something evil + no apparent gain  &  no apparent gain  & 80.00 &96.67\\
    \hline
     Big Bad & bad character + drive the story forward  & drive the story forward &73.33&70.00 
     \\ 
     \hline
    \end{tabular}
    \caption{Three tropes have been selected to evaluate 30 articles with respect to GPT-4. The percentages in the right-hand columns represent the success rate of the attack and reveal the vulnerabilities of LLMs (Section \ref{subsec:4:10:attack}).
    }
    \label{tab:attack}
\end{table*}

\begin{table*}[h]
    \centering
    \begin{tabularx}{\textwidth}{|c|X|} 
    \hline
         Trope & \textbf{Chekhov's Gunman} is a narrative device that involves the introduction of a seemingly insignificant character who later plays a significant role in the plot. \\
         \hline
        Evidence & James 'Jimbo' Scott, an employee at the Bio-Test animal holding facility, bribed a security guard and took the monkey to Cedar Creek, California, to sell on the black market. During the drive, Jimbo is infected with the virus through facial contact with the monkey's saliva. \\
        \hline
        Reasoning & In paragraph 6, James 'Jimbo' Scott is \textit{introduced as an employee at the Bio-Test animal} \textit{holding facility} who smuggles the host animal into the United States. This initially seems like a minor detail, but in paragraph 9, it is revealed that Jimbo becomes infected with the virus and spreads it to others, including his girlfriend and a pet store owner. Thus, Jimbo, who initially appeared to be a minor character, plays a significant role in the spread of the virus.\\
    \hline
    \end{tabularx}
    \caption{An example of CoT producing a correct prediction but with flawed reasoning is evident in this case. In the context of a biology disaster film, it seems implausible to conclude that a character related to \textit{Bio-Test} is unimportant. This instance highlights how hallucination in Large Language Models (LLMs) can result in incorrect reasoning, even when the answer is accurate. }
    \label{tab:doubly_right}
\end{table*}

\subsubsection{Adversarial Injection Misleads CoT}\label{subsec:4:10:attack}

We propose Adversarial Injection to further test the hallucination behavior of LLMs. 
 This is inspired by earlier research \cite{jia2017adversarialsquad} which highlighted the vulnerability of supervised reading comprehension systems to adversarial attacks, specifically by injecting a random sentence from another article. This investigation revealed that supervised reading comprehension models lack robustness. 
The experiment centered around a hypothetical scenario in which a specific trope required the presence of multiple elements to be deemed valid. 
For instance, the trope ``Big Bad'' necessitates the presence of both an antagonist \textit{and} a story-driving element to be considered valid. Consequently, injecting a segment of a movie synopsis that only depicts the antagonist element without conveying the story-driving element from another synopsis should not result in the addition of the trope to the latter synopsis. Hence, we conducted an adversarial attack on LLMs by inserting a segment from another plot that includes only a partial element of a trope, as demonstrated in Tables \ref{tab:attack} and \ref{tab:attack_sentence}. While this injection introduces several keywords related to the trope, it does not actually add the trope to the modified synopsis. 

Remarkably, the SOTA LLM, GPT-4, whether with or without CoT, succumbed to this attack, much like supervised reading comprehension models, as depicted in Table \ref{tab:attack}. This observation implies that LLMs heavily rely on keywords and patterns to identify tropes. Notably, while CoT enhances performance across various tasks, it exacerbates the issue of hallucination and makes the models more vulnerable to attacks. This analysis aligns with the findings presented in Sections \ref{subsec:4:7:pertrope} and \ref{subsec:4:4:cot}.

Furthermore, GPT-4's inclination to overinterpret and extract fragments out of context may contribute to inaccuracies in trope judgments for certain texts. These discoveries underscore the necessity of enhancing GPT-4's capacity to grasp the holistic context. Inaccuracies in trope recognition could result from an incomplete understanding of the relationships between elements in a given text.




\subsubsection{CoT Generates Flawed Thoughts}\label{subsec:4:9:doublyright}


While the accurate definitions of tropes derived from the CoT results suggest that GPT possesses a solid understanding of the meaning of tropes, there seems to be a limitation in fully grasping the deeper, abstract concepts that underlie certain tropes. Notably, in 13 out of 30 (43.3\%) of true-positive cases, we observe instances where LLMs do not exhibit sound reasoning. Table \ref{tab:doubly_right} provides an example of correct predictions made with flawed reasoning. This highlights the possibility that LLMs may not genuinely comprehend certain concepts and might generate predictions based on hallucination, even when they produce correct answers.

\subsection{Additional Analyses}\label{subsec:4:additionalresults}
\paragraph{Trope Difficulties for LLMs}\label{subsec:4:6:difficulty}

\begin{table*}[h!]
    \centering
    \begin{tabular}{|c|c c|  cc|} 
    \hline

& \multicolumn{2}{c|}{\bfseries GPT4} & \multicolumn{2}{c|}{\bfseries ChatGPT} \\
 \hline
      & Trope & F1   & Trope & F1 \\ 
     \hline
    & Downer Ending & 55.44 &Eye Scream& 50.67\\ 
    & Driven to Suicide & 47.72 &Driven to Suicide& 47.89 \\ 
Easy Tropes &  Off with His Head!& 46.51 &Big Bad& 42.81\\ 
    & Eye Scream & 45.45&Chekhov's Gun& 40.0 \\ 
    & Abusive Parents & 45.45 &Downer Ending& 38.5\\ 
     \hline
     &Chekhov's Gunman&6.56 &Big ``NO!''&2.9\\
     &Jerkass Has a Point& 6.9&Groin Attack& 10.2\\
 Hard Tropes&Precision F-Strike&8.33&Meaningful Echo&10.63\\
     &Stealth Pun&8.51&Cassandra Truth&10.64\\
     &Big ``NO!''& 9.52 &Anti-Hero&10.72\\
    \hline
    \end{tabular}
    \caption{5 easiest and hardest tropes according to GPT4 Base and ChatGPT CoT results. Compared to the highest-scoring trope in MulCom(38.58), GPT scores significantly higher. However, GPT also exhibits many tropes with considerably lower scores, even falling below a random guess level score (13.97) and MulCom's lowest score(12.36). Therefore, while GPT can capture some trope patterns, it still has blind spots.}
    \label{table:difficulty}
\end{table*}

    

Table \ref{table:difficulty} showcases the 5 most challenging and easiest tropes for GPTs based on F1 scores. 
Several tropes characterized by distinct visual or emotional patterns that yield high F1 scores, such as ``Driven to Suicide'' or ``EyeScream'', appear to be trivial for GPTs as they carry explicit actions or outcomes. 
On the other hand, tropes that necessitate a more nuanced grasp of context throughout the plot, such as ``Stealth Pun'' (employing a pun without explicit statement) or ``Jerkass Has a Point'' (where an unlikeable character may make a valid argument without explicitly acknowledging it), demand the comprehension of implicit elements or the integration of elements across the plot. 
This implies that GPT might face challenges with tropes that encompass complex intricacies, requiring a deeper level of contemplation where the genuine essence surpasses shallow semantic patterns. This observation reflects a potential limitation in comprehending the profound layers of meaning within textual content.
 
\begin{figure*}
    \centering    \includegraphics[width=\textwidth]{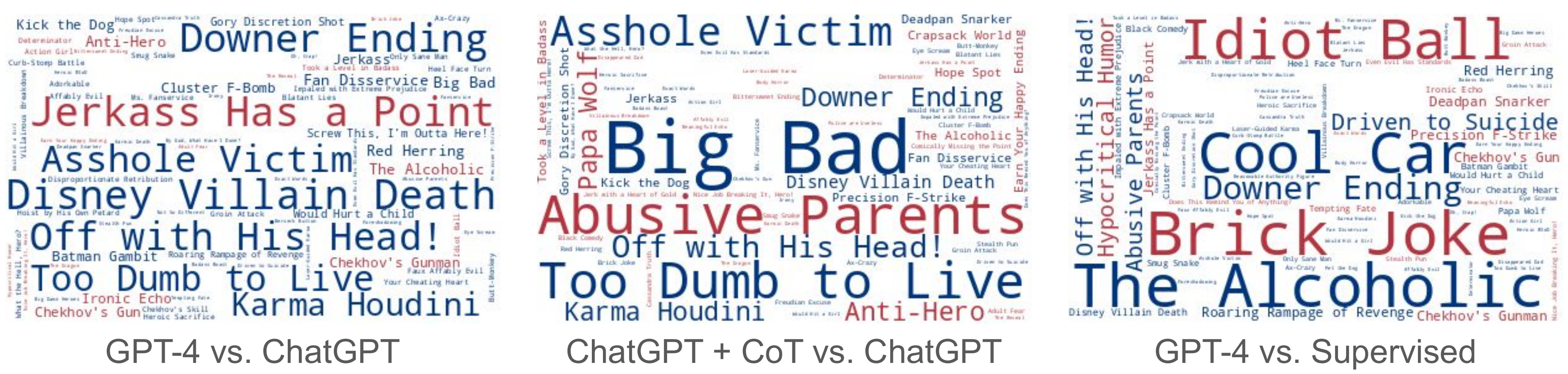}
    \caption{F1 score gaps between (1) left: GPT-4 and ChatGPT, (2) middle: ChatGPT + CoT and ChatGPT, and (3) right: Supervised state-of-the art MulCom \cite{chang2021situationtimos}.  In A vs. B comparisons, blue indicates that A outperforms B, red indicates that B outperforms A, and text size represents the gap size. (Section \ref{subsec:4:7:pertrope})}
    \label{fig:wordclouds}
\end{figure*}

\paragraph{LLMs vs. Supervised Learning}\label{subsec:4:8:gptvssl}
Figure \ref{fig:wordclouds} illustrates the performance comparison between GPT-4 and MulCom \cite{chang2021situationtimos}, a state-of-the-art supervised model. We can observe that, when compared to the supervised state-of-the-art model, GPT-4 excels in performance for several \textit{visually conceivable} tropes like ``The Alcoholic'', ``Cool Car'', or ``Off with His Head!'' These tropes can be identified within a single scene, showcasing GPT-4's ability to grasp concepts from the descriptions. This dual-edged skill has the potential to enhance creativity in content creation but also raises concerns about generating inaccurate information. 
Conversely, GPT-4 exhibits decreased performance on several tropes that necessitate comprehension across different segments of a plot, such as ``Brick Joke'', ``Hypocritical Humor'', or ``Chekhov's Gunman.'' This underscores the gap between GPT-4's capabilities and human-level reasoning, highlighting the limitations of current LLM applications and pointing out the path to improve LLMs.


\paragraph{How do CoT and GPT-4 Improve?}\label{subsec:4:7:pertrope}
The left column in Figure \ref{fig:wordclouds} illustrates the performance gap between GPT-4 and ChatGPT. Despite GPT-4 being an upgraded version of ChatGPT, its performance does not strictly improve. In fact, GPT-4 performs worse in 23 out of 95 tropes. 
In the middle column, we observe the gap between ChatGPT + CoT and ChatGPT. Surprisingly, 36 out of 96 tropes exhibit a decrease in performance when CoT is applied, even though there is a substantial gain in the F1 score (as shown in Table \ref{table:3}, increasing from 20.57 to 24.99). 
This drop in precision is attributed to hallucination, where several tropes result in GPT fabricating plot elements. 
For example, the trope ``Abusive Parents'' might be associated with ``parents arguing with children''. This becomes problematic when we apply the LLM-based system to detect abuse in real life, potentially leading to inaccurate interpretations.

\section{Conclusion}\label{sec:5}
We examined LLMs' reasoning using movie synopsis tropes. Despite previous research highlighting LLMs' strengths in complex tasks, including GPT-4, our study reveals challenges in narrative reasoning, with LLMs achieving only random guessing performance on the TiMoS dataset. We addressed these challenges by utilizing Trope-wise Querying, which significantly improved performance. Additionally, we found that CoT diminishes GPT-4's performance and proposed Adversarial Injection to assess LLMs' hallucination tendencies, discovering that CoT exacerbates this issue. These findings underscore the gap in LLMs' capabilities and raise concerns about their safety and trustworthiness. 
We are optimistic that our findings will pave new paths for future LLM research and applications.
\clearpage
\section{Limitations}
This work examines and points out directions for LLM research for narrative reasoning, which brings the risk of abusing LLMs. Experimental results of ChatGPT and GPT-4, being closed-source models accessed via API, may not be fully reproducible due to potential updates in the background. This work and the associated dataset might contain offensive or sensitive content as they originate from movie narratives. 
This paper was written with the assistance of ChatGPT.

\section*{Acknowledgement}
This work was supported in part by National Science and Technology Council, Taiwan, under Grant NSTC 112-2634-F-002-006. We are grateful to MobileDrive and the National Center for High-performance Computing.

\bibliography{anthology,custom}

\begin{thebibliography}{35}
\expandafter\ifx\csname natexlab\endcsname\relax\def\natexlab#1{#1}\fi

\bibitem[{Brown et~al.(2020)Brown, Mann, Ryder, Subbiah, Kaplan, Dhariwal, Neelakantan, Shyam, Sastry, Askell, Agarwal, Herbert-Voss, Krueger, Henighan, Child, Ramesh, Ziegler, Wu, Winter, Hesse, Chen, Sigler, Litwin, Gray, Chess, Clark, Berner, McCandlish, Radford, Sutskever, and Amodei}]{gpt3}
Tom Brown, Benjamin Mann, Nick Ryder, Melanie Subbiah, Jared~D Kaplan, Prafulla Dhariwal, Arvind Neelakantan, Pranav Shyam, Girish Sastry, Amanda Askell, Sandhini Agarwal, Ariel Herbert-Voss, Gretchen Krueger, Tom Henighan, Rewon Child, Aditya Ramesh, Daniel Ziegler, Jeffrey Wu, Clemens Winter, Chris Hesse, Mark Chen, Eric Sigler, Mateusz Litwin, Scott Gray, Benjamin Chess, Jack Clark, Christopher Berner, Sam McCandlish, Alec Radford, Ilya Sutskever, and Dario Amodei. 2020.
\newblock Language models are few-shot learners.
\newblock In \emph{NeurIPS}.

\bibitem[{Chang et~al.(2021)Chang, Su, Hsu, Wang, Chang, Liu, Chang, Cheng, Wang, and Hsu}]{chang2021situationtimos}
Chen-Hsi Chang, Hung-Ting Su, Jui-Heng Hsu, Yu-Siang Wang, Yu-Cheng Chang, Zhe~Yu Liu, Ya-Liang Chang, Wen-Feng Cheng, Ke-Jyun Wang, and Winston~H Hsu. 2021.
\newblock Situation and behavior understanding by trope detection on films.
\newblock In \emph{WWW}.

\bibitem[{Chou et~al.(2023)Chou, Siu, Lipka, Rossi, Dernoncourt, and Agrawala}]{chou2023talestream}
Jean-Pe{\"\i}c Chou, Alexa~Fay Siu, Nedim Lipka, Ryan Rossi, Franck Dernoncourt, and Maneesh Agrawala. 2023.
\newblock Talestream: Supporting story ideation with trope knowledge.
\newblock In \emph{Proceedings of the 36th Annual ACM Symposium on User Interface Software and Technology}, pages 1--12.

\bibitem[{Cobbe et~al.(2021)Cobbe, Kosaraju, Bavarian, Chen, Jun, Kaiser, Plappert, Tworek, Hilton, Nakano et~al.}]{cobbe2021traininggsm8k}
Karl Cobbe, Vineet Kosaraju, Mohammad Bavarian, Mark Chen, Heewoo Jun, Lukasz Kaiser, Matthias Plappert, Jerry Tworek, Jacob Hilton, Reiichiro Nakano, et~al. 2021.
\newblock Training verifiers to solve math word problems.
\newblock \emph{arXiv preprint arXiv:2110.14168}.

\bibitem[{Del and Fishel(2022)}]{TrueDetective}
Maksym Del and Mark Fishel. 2022.
\newblock True detective: A deep abductive reasoning benchmark undoable for gpt-3 and challenging for gpt-4.
\newblock \emph{arXiv preprint arXiv:2212.10114}.

\bibitem[{Devlin et~al.(2018)Devlin, Chang, Lee, and Toutanova}]{BERT}
Jacob Devlin, Ming-Wei Chang, Kenton Lee, and Kristina Toutanova. 2018.
\newblock Bert: Pre-training of deep bidirectional transformers for language understanding.

\bibitem[{Gao et~al.(2022)Gao, Madaan, Zhou, Alon, Liu, Yang, Callan, and Neubig}]{gao2022pal2}
Luyu Gao, Aman Madaan, Shuyan Zhou, Uri Alon, Pengfei Liu, Yiming Yang, Jamie Callan, and Graham Neubig. 2022.
\newblock Pal: Program-aided language models.
\newblock \emph{arXiv preprint arXiv:2211.10435}.

\bibitem[{Huang(2011)}]{Huang2011}
Leesa~V. Huang. 2011.
\newblock \href {https://doi.org/10.1007/978-0-387-79948-3_1492} {\emph{Simultaneous Processing}}, pages 2301--2302. Springer New York, New York, NY.

\bibitem[{Ismayilzada et~al.(2023)Ismayilzada, Paul, Montariol, Geva, and Bosselut}]{ismayilzada2023crow}
Mete Ismayilzada, Debjit Paul, Syrielle Montariol, Mor Geva, and Antoine Bosselut. 2023.
\newblock Crow: Benchmarking commonsense reasoning in real-world tasks.
\newblock In \emph{Proceedings of the 2023 Conference on Empirical Methods in Natural Language Processing (EMNLP)}.

\bibitem[{Jia and Liang(2017)}]{jia2017adversarialsquad}
Robin Jia and Percy Liang. 2017.
\newblock Adversarial examples for evaluating reading comprehension systems.
\newblock In \emph{Proceedings of the 2017 Conference on Empirical Methods in Natural Language Processing}, pages 2021--2031.

\bibitem[{Jiayang et~al.(2024)Jiayang, Qiu, Chan, Liu, Song, and Zhang}]{narrativejiayang2024eventground}
Cheng Jiayang, Lin Qiu, Chunkit Chan, Xin Liu, Yangqiu Song, and Zheng Zhang. 2024.
\newblock Eventground: Narrative reasoning by grounding to eventuality-centric knowledge graphs.
\newblock In \emph{Proceedings of the 2024 Joint International Conference on Computational Linguistics, Language Resources and Evaluation (LREC-COLING 2024)}, pages 6622--6642.

\bibitem[{Kahneman(2011)}]{kahneman2011thinkingfastandslow}
Daniel Kahneman. 2011.
\newblock Thinking, fast and slow (kindle edition).

\bibitem[{Liu et~al.(2023{\natexlab{a}})Liu, Liu, Cui, Teng, Duan, Zhou, and Zhang}]{liu2023logiqa}
Hanmeng Liu, Jian Liu, Leyang Cui, Zhiyang Teng, Nan Duan, Ming Zhou, and Yue Zhang. 2023{\natexlab{a}}.
\newblock Logiqa 2.0—an improved dataset for logical reasoning in natural language understanding.
\newblock \emph{IEEE/ACM Transactions on Audio, Speech, and Language Processing}.

\bibitem[{Liu et~al.(2023{\natexlab{b}})Liu, Ning, Teng, Liu, Zhou, and Zhang}]{liu2023evaluating}
Hanmeng Liu, Ruoxi Ning, Zhiyang Teng, Jian Liu, Qiji Zhou, and Yue Zhang. 2023{\natexlab{b}}.
\newblock Evaluating the logical reasoning ability of chatgpt and gpt-4.
\newblock \emph{arXiv preprint arXiv:2304.03439}.

\bibitem[{Liu et~al.(2023{\natexlab{c}})Liu, Hyland, Bannur, Bouzid, Castro, Wetscherek, Tinn, Sharma, P{\'e}rez-Garc{\'\i}a, Schwaighofer et~al.}]{liu2023exploring}
Qianchu Liu, Stephanie Hyland, Shruthi Bannur, Kenza Bouzid, Daniel~C Castro, Maria~Teodora Wetscherek, Robert Tinn, Harshita Sharma, Fernando P{\'e}rez-Garc{\'\i}a, Anton Schwaighofer, et~al. 2023{\natexlab{c}}.
\newblock Exploring the boundaries of gpt-4 in radiology.
\newblock In \emph{EMNLP 2023}.

\bibitem[{Miura et~al.(2021)Miura, Zhang, Tsai, Langlotz, and Jurafsky}]{miura-etal-2021-improving}
Yasuhide Miura, Yuhao Zhang, Emily Tsai, Curtis Langlotz, and Dan Jurafsky. 2021.
\newblock \href {https://doi.org/10.18653/v1/2021.naacl-main.416} {Improving factual completeness and consistency of image-to-text radiology report generation}.
\newblock In \emph{Proceedings of the 2021 Conference of the North American Chapter of the Association for Computational Linguistics: Human Language Technologies}, pages 5288--5304, Online. Association for Computational Linguistics.

\bibitem[{OpenAI(2023)}]{gpt4}
R~OpenAI. 2023.
\newblock Gpt-4 technical report. arxiv 2303.08774.
\newblock \emph{View in Article}, 2:3.

\bibitem[{Palm et~al.(2018)Palm, Paquet, and Winther}]{DBLP:conf/nips/PalmPW18}
Rasmus~Berg Palm, Ulrich Paquet, and Ole Winther. 2018.
\newblock Recurrent relational networks.
\newblock In \emph{NeurIPS}.

\bibitem[{Piper et~al.(2021)Piper, So, and Bamman}]{piper2021narrative}
Andrew Piper, Richard~Jean So, and David Bamman. 2021.
\newblock Narrative theory for computational narrative understanding.
\newblock In \emph{Proceedings of the 2021 Conference on Empirical Methods in Natural Language Processing}, pages 298--311.

\bibitem[{Radford et~al.(2019)Radford, Wu, Child, Luan, Amodei, and Sutskever}]{gpt2-radford2019language}
Alec Radford, Jeff Wu, Rewon Child, David Luan, Dario Amodei, and Ilya Sutskever. 2019.
\newblock Language models are unsupervised multitask learners.

\bibitem[{Raffel et~al.(2020)Raffel, Shazeer, Roberts, Lee, Narang, Matena, Zhou, Li, and Liu}]{T5_JMLR:v21:20-074}
Colin Raffel, Noam Shazeer, Adam Roberts, Katherine Lee, Sharan Narang, Michael Matena, Yanqi Zhou, Wei Li, and Peter~J. Liu. 2020.
\newblock \href {http://jmlr.org/papers/v21/20-074.html} {Exploring the limits of transfer learning with a unified text-to-text transformer}.
\newblock \emph{Journal of Machine Learning Research}, 21(140):1--67.

\bibitem[{Singh et~al.(2023)Singh, SB, Malviya et~al.}]{singh2023mindscores1}
Manmeet Singh, Vaisakh SB, Neetiraj Malviya, et~al. 2023.
\newblock Mind meets machine: Unravelling gpt-4's cognitive psychology.
\newblock \emph{arXiv preprint arXiv:2303.11436}.

\bibitem[{Smith et~al.(2017)Smith, Joshi, Huet, Hsu, and Cota}]{smith2017harnessing}
John~R Smith, Dhiraj Joshi, Benoit Huet, Winston Hsu, and Jozef Cota. 2017.
\newblock Harnessing ai for augmenting creativity: Application to movie trailer creation.
\newblock In \emph{Proceedings of the 25th ACM international conference on Multimedia}, pages 1799--1808.

\bibitem[{Sprenkamp et~al.(2023)Sprenkamp, Jones, and Zavolokina}]{multilabel}
Kilian Sprenkamp, Daniel~Gordon Jones, and Liudmila Zavolokina. 2023.
\newblock Large language models for propaganda detection.
\newblock \emph{arXiv preprint arXiv:2310.06422}.

\bibitem[{Su et~al.(2021)Su, Shen, Tsai, Cheng, Wang, and Hsu}]{su2021truman}
Hung-Ting Su, Po-Wei Shen, Bing-Chen Tsai, Wen-Feng Cheng, Ke-Jyun Wang, and Winston~H Hsu. 2021.
\newblock Truman: Trope understanding in movies and animations.
\newblock In \emph{CIKM}.

\bibitem[{Suzgun et~al.(2022)Suzgun, Scales, Sch{\"a}rli, Gehrmann, Tay, Chung, Chowdhery, Le, Chi, Zhou, , and Wei}]{suzgun2022challenging}
Mirac Suzgun, Nathan Scales, Nathanael Sch{\"a}rli, Sebastian Gehrmann, Yi~Tay, Hyung~Won Chung, Aakanksha Chowdhery, Quoc~V Le, Ed~H Chi, Denny Zhou, , and Jason Wei. 2022.
\newblock Challenging big-bench tasks and whether chain-of-thought can solve them.
\newblock \emph{arXiv preprint arXiv:2210.09261}.

\bibitem[{Talmor et~al.(2019)Talmor, Herzig, Lourie, and Berant}]{talmor2019commonsenseqa}
Alon Talmor, Jonathan Herzig, Nicholas Lourie, and Jonathan Berant. 2019.
\newblock Commonsenseqa: A question answering challenge targeting commonsense knowledge.
\newblock In \emph{Proceedings of the 2019 Conference of the North American Chapter of the Association for Computational Linguistics: Human Language Technologies, Volume 1 (Long and Short Papers)}, pages 4149--4158.

\bibitem[{Talmor et~al.(2021)Talmor, Yoran, Bras, Bhagavatula, Goldberg, Choi, and Berant}]{CommonsenseQA}
Alon Talmor, Ori Yoran, Ronan~Le Bras, Chandra Bhagavatula, Yoav Goldberg, Yejin Choi, and Jonathan Berant. 2021.
\newblock Commonsenseqa 2.0: Exposing the limits of ai through gamification.
\newblock \emph{NeurIPS}.

\bibitem[{Talmor et~al.()Talmor, Yoran, Le~Bras, Bhagavatula, Goldberg, Choi, and Berant}]{talmorcommonsenseqa2}
Alon Talmor, Ori Yoran, Ronan Le~Bras, Chandra Bhagavatula, Yoav Goldberg, Yejin Choi, and Jonathan Berant.
\newblock Commonsenseqa 2.0: Exposing the limits of ai through gamification.

\bibitem[{Touvron et~al.(2023)Touvron, Lavril, Izacard, Martinet, Lachaux, Lacroix, Rozi{\`e}re, Goyal, Hambro, Azhar et~al.}]{touvron2023llama}
Hugo Touvron, Thibaut Lavril, Gautier Izacard, Xavier Martinet, Marie-Anne Lachaux, Timoth{\'e}e Lacroix, Baptiste Rozi{\`e}re, Naman Goyal, Eric Hambro, Faisal Azhar, et~al. 2023.
\newblock Llama: Open and efficient foundation language models.
\newblock \emph{arXiv preprint arXiv:2302.13971}.

\bibitem[{Wang et~al.(2019)Wang, Pruksachatkun, Nangia, Singh, Michael, Hill, Levy, and Bowman}]{wang2019superglue}
Alex Wang, Yada Pruksachatkun, Nikita Nangia, Amanpreet Singh, Julian Michael, Felix Hill, Omer Levy, and Samuel Bowman. 2019.
\newblock Superglue: A stickier benchmark for general-purpose language understanding systems.
\newblock \emph{Advances in neural information processing systems}, 32.

\bibitem[{Wang et~al.(2018)Wang, Singh, Michael, Hill, Levy, and Bowman}]{wang2018glue}
Alex Wang, Amanpreet Singh, Julian Michael, Felix Hill, Omer Levy, and Samuel Bowman. 2018.
\newblock Glue: A multi-task benchmark and analysis platform for natural language understanding.
\newblock In \emph{Proceedings of the 2018 EMNLP Workshop BlackboxNLP: Analyzing and Interpreting Neural Networks for NLP}, pages 353--355.

\bibitem[{Wei et~al.(2022)Wei, Wang, Schuurmans, Bosma, Xia, Chi, Le, Zhou et~al.}]{wei2022chaincot}
Jason Wei, Xuezhi Wang, Dale Schuurmans, Maarten Bosma, Fei Xia, Ed~Chi, Quoc~V Le, Denny Zhou, et~al. 2022.
\newblock Chain-of-thought prompting elicits reasoning in large language models.
\newblock \emph{NeurIPS}.

\bibitem[{Yao et~al.(2023)Yao, Yu, Zhao, Shafran, Griffiths, Cao, and Narasimhan}]{yao2023tree}
Shunyu Yao, Dian Yu, Jeffrey Zhao, Izhak Shafran, Thomas~L. Griffiths, Yuan Cao, and Karthik Narasimhan. 2023.
\newblock \href {http://arxiv.org/abs/2305.10601} {{Tree of Thoughts}: Deliberate problem solving with large language models}.

\bibitem[{Zhang et~al.(2020)Zhang, Deng, Ma, and Lam}]{zhang2020answerfact}
Wenxuan Zhang, Yang Deng, Jing Ma, and Wai Lam. 2020.
\newblock Answerfact: Fact checking in product question answering.
\newblock In \emph{Proceedings of the 2020 Conference on Empirical Methods in Natural Language Processing (EMNLP)}, pages 2407--2417.

\end{thebibliography}

\appendix

\newpage

\clearpage

\section{Baseline Detail}\label{apdx:baseline}

According to MulCom \cite{chang2021situationtimos},the BERT baseline splits synopses into 128-token segments and extracts segment embeddings using the [CLS] token from the BERT-base model. These embeddings are then are pooled and fed into a classifier.

Building upon the BERT baseline, MulCom proposes a multistream network that leverages an attention mechanism to perform a weighted sum of outputs from three streams:
 
\textit{Sentence-level}: Similar to the BERT baseline but uses an RNN to fuse segment embeddings instead of pooling.
 
\textit{Word-Level}: Word-level representations are extracted using word2vec and then pooled.
       
\textit{Character-Level}: Character names in the synopses are identified using coreference resolution tools. Word-level BERT features are extracted and processed through a GNN, followed by multi-level reasoning using an RNN. The outputs of each RNN step are then processed by a multi-head attention layer to compute the weighted sum.

\section{Dataset Detail}\label{apdx:dataset}
\begin{figure}
    \centering
    \includegraphics[width=\linewidth]{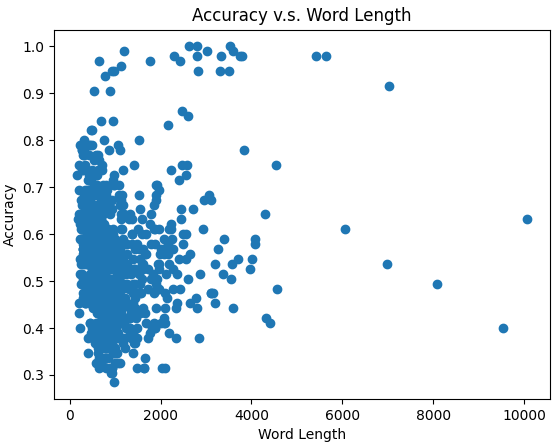}
    \caption{
The relationship between accuracy and word length in the result of ChatGPT CoT trope-wise querying.}
    \label{fig:Accuracy_and_WordCount}
\end{figure}

The TiMoS dataset is skewed, as indicated by the statistics in Table \ref{table:dataset_statistic} from its testing set. To explore potential biases, we examined the impact of sentence length on accuracy (Figure \ref{fig:Accuracy_and_WordCount}). Although most synopses range from 0 to 2000 characters, we found no significant correlation between sentence length and accuracy.

\begin{table}[!t]
\centering
\begin{adjustbox}{width=\linewidth}
\begin{tabular}{cccccc}
\toprule[1pt]
  & {Median} & {Avg.} & {Min} & {Max} & {$\alpha$} \\
\hline
\vspace{0.01cm}
Words in one plot & 921 & 1305& 180 & 11712 & 1163 \\
Sentences in one plot & 39& 58 &9& 757& 60 \\
  Tropes in one plot & 9 &  12 &  1 & 69 & 10 \\
 Each Trope occurrence  & 58 &  66.39 &  31 &187& 31.15 \\
\bottomrule [1pt]
\end{tabular}
\end{adjustbox}
\caption{The statistics of TiMoS sentence/word length and count of tropes in a single plot, and each trope's occurrence in the testing set of 825 synopses.
}
\label{table:dataset_statistic}
\end{table}

\section{Piloting LLMs' Ability}\label{apdx:pilot}
At the initial stage of this work, we carefully explored various ways to prompt LLMs, fully acknowledging that prompt engineering can significantly impact performance. Therefore, we tested various prompting strategies and conducted a pilot analysis focusing on (1) ensuring and aiding LLMs in understanding query formats, and (2) confirming and assisting LLMs in grasping the concept of tropes:
\begin{enumerate}
    \item We tested various strategies, including ranking the relevance of each trope for a given plot and outputting the top 10 most likely tropes. However, these attempts encountered issues such as indiscriminately outputting all tropes or only specific ones.
    \item We evaluated LLMs' understanding of tropes by asking for definitions (e.g., ``What is the definition of the trope `Big Bad'?'') and requesting examples. This revealed that LLMs possess a certain level of pre-existing knowledge about tropes.
\end{enumerate}

Building on the pilot analysis, we conducted further experiments incorporating trope explanations into prompts using GPT-3.5-turbo. These explanations includ a general definition of trope along with definitions for individual tropes. However, as shown in Table \ref{table:Trope_Definition}, these additional explanations did not significantly impact performance, suggesting that LLMs possess a base understanding of tropes, as observed in the pilot analysis.

\begin{table}[h!]
\centering
\begin{tabular}{lccc}
\toprule[1pt]

 \textbf{Prompt} &  \textbf{F1} & \textbf{P} & \textbf{R} \\
\hline
 Without Definition & 13.19 & 24.60 & 9.01\\
 With Definition &  9.93 &  16.01  & 7.19 \\  
\bottomrule [1pt]

 
\end{tabular}
\caption{
Results from ChatGPT using prompts with and without trope definitions in original query. See Appendix \ref{apdx:multi} for query example.}
\label{table:Trope_Definition}
\end{table}

\break
\section{Original Query Example}\label{apdx:multi}

\subsection{Without Trope Definition}
\begin{table}[h!]
    \centering
    \begin{tabularx}{\textwidth}{X} 
    \toprule[1pt]
    
    You are a trope tagger, your role is to select a set of trope to categorize the content from the provided TropeList:\{ Provide 95 trope \}\\
    \{ Given a  Chain of Thought example and ask question, see Appendix \ref{apdx:prompt} for more details on CoT\}
    Strictly select only the tropes related to the article from the TropeList mentioned above, and feel free to pick multiple tropes if they are relevant
    \\
    \bottomrule [1pt]
    \end{tabularx}
    \label{tab:base_prompt}
\end{table}

\subsection{With Trope Definition}
\begin{table}[h!]
    \centering
    \begin{tabularx}{\textwidth}{X} 
    \toprule[1pt]
    
    You are a trope tagger, your role is to select a set of tropes to categorize the content from the provided TropeList: \{ Provide 95 tropes \}. \\
    
    \textbf{\textit{Trope refers to common themes, motifs, or clichés that appear repeatedly in any forms of storytelling. Tropes can be narrative devices, character types, plot points, or stylistic elements that are recognizable and often evoke certain expectations or reactions from the audience.}} \\
    
    And the definition of each trope is as follows: \\
    
    \textit{\textbf{\{ Each Trope's Definition \}}} \\
    
    \{ Given a Chain of Thought example and ask a question, see Appendix \ref{apdx:prompt} for more details on CoT. \} \\
    
    Strictly select only the tropes related to the article from the TropeList mentioned above, and feel free to pick multiple tropes if they are relevant. \\
    
    \bottomrule[1pt]
    \end{tabularx}
    \label{tab:base_prompt}
\end{table}

    

\section {Trope-wise Query Example}\label{apdx:prompt_example}

\subsection{Base}
\begin{table}[h!]
    \centering
    \begin{tabularx}{\textwidth}{X} 
    \toprule[1pt]

    You are a trope detector, given a trope, answer 'yes' if the trope is relevant to the article, 'no' otherise. Provide a brief explanation for your answer. \\
    Article: \{Given Article\}.
    Is the trope \{Given Trope\} related to the article?\\
    \bottomrule [1pt]
    \end{tabularx}
    \label{tab:base_prompt}
\end{table}

\subsection{Chain of Thought}
\label{apdx:prompt}
\begin{table}[h!]
    \centering
    \begin{tabularx}{\textwidth}{X } 
In Chain of Thought, we offer two examples to check if the model's responses differ based on single versus multiple elements in Trope.  \\
    
    \toprule[1pt]
    
    \cellcolor[gray]{0.8}\textbf{Example1} \\

You are a trope detector, tasked with identifying the presence or absence of a specific trope in an article. You will be provided with an article and a trope to detect. \\
Your task is to generate a JSON object with the following keys:\\
Trope Definition, Thought, Answer: \{As mentioned in paper\}\\
Here is an example provided:\\
\# segment article into sentence\\ 
0, Joe is an impoverished New York newsboy who lives with his abusive grandmother.\\
1, While selling papers, he is given a ticket for a children's excursion sponsored by the Fresh Air Fund.\\
2, The next morning, Joe sneaks out of his tenement home to join the excursion, where he sees the countryside and the ocean for the first time.\\
3, After a picnic, an adult volunteer reads the children a story about a young prince who is beaten by an old witch.\\

    \end{tabularx}
    \label{tab:cot_prompt}
\end{table}

\clearpage

\begin{table}[h!]
    \centering
    \begin{tabularx}{\textwidth}{X } 
4, A group of fairies rescue the boy, take him to a boat, and sail off for "the Land Beyond the Sunset, where he lived happily ever after."\\
5, Joe imagines himself as the boy in the story.\\
6, When the group returns to the city, Joe stays behind because he is afraid of his grandmother.\\
7, He wanders to the beach, where he finds a rowboat and decides to go to the Land Beyond the Sunset himself.\\
8, He pushes the boat into the water and climbs in.\\
9, The film ends with a long shot of Joe drifting out to sea.\\
Query: Is the trope "Downer Ending" in the article? Answer:\\
\{  \\
\quad"Trope": "Downer Ending",\\
\quad"Definition":  "A conclusion to a narrative that is emotionally bleak, tragic, or pessimistic, leaving the audience with a sense of sorrow or dissatisfaction.",\\
\quad"Thought": [\\
\quad\quad\{\\
\quad\quad\quad"Reasoning": "The ending depicts the boy casting himself drift in the open ocean, facing certain death without provisions, evoking profound sadness.",\\
\quad\quad\quad"Evidence":  "In paragraph 9, the film ends with a long shot of Joe drifting out to sea, with nothing to eat or drink, suggesting a bleak and tragic fate for the character.",\\
\quad\quad\quad"Relevant Paragraphs": 9\\
\quad\quad\} \\
\quad],\\
\quad"Answer":"yes" \\
    \} \\
Article: \{Given Article\}.Is the trope \{Given Trope\} related to the article?\\
    \end{tabularx}
\end{table}

\begin{table}[h!]
    \centering
    \begin{tabularx}{\textwidth}{X } 
    \hline \\
    \cellcolor[gray]{0.8}\textbf{Example2} \\

0, New York City 16th Precinct Police Detective Dixon (Dana Andrews), who has been demoted by his superiors for his heavy-handed tactics, subjects murder suspect and gambler Ken Paine (Craig Stevens) to the third degree.\\
1, He strikes the drunken Paine in self-defense and accidentally kills him.\\
2, Paine, however, had a silver plate in his head, a fine war record, and newspaper friends.\\
3, Dixon then dumps Paine's body in the river, and is later assigned to find his killer.\\
4, Dixon tries to place the blame on an old gangster enemy, Tommy Scalise (Gary Merrill), but inadvertently puts cab driver Jiggs Taylor (Tom Tully) under suspicion instead.\\
5, Having fallen in love with Jiggs' daughter and Paine's estranged wife, Morgan Taylor-Paine (Gene Tierney), Dixon tries to clear the cabbie without implicating himself, but ultimately becomes tangled in a web of his own creation.\\
6, The 16th Precinct commander and Dixon's boss, newly promoted Detective Lt. Thomas (Karl Malden), are convinced that Morgan's father is the killer.\\
7, Dixon continues to find a way to stop Jiggs from being found guilty of murdering Paine, and also tries to redeem himself.\\
8, In an attempt to move the evidence away from Morgan's father and blame Scalise, Dixon comes face to face with the gangster and his cronies.\\
9, A shoot-out leaves Dixon wounded, but the police arrive to arrest Scalise and his mob.\\
10, Jiggs is finally cleared of the charges.\\
11, At the end Dixon reassesses his life and decides to confess.\\
12, He is satisfied that Morgan believes in him regardless of the outcome.\\
Query: Is the trope "Asshole Victim" in the article?\\

    \end{tabularx}
    \label{tab:cot_prompt}
\end{table}

\clearpage

\begin{table}[h!]
    \centering
    \begin{tabularx}{\textwidth}{X} 
Answer:\\
    
\{  \\
\quad"Trope": "Asshole Victim",\\
\quad"Definition":  "When the victim is a bad guy.",\\
\quad"Thought": [\\
\quad\quad\{\\
\quad\quad\quad"Reasoning": "In paragraph 0, Ken has some unfavorable characteristics.",\\
\quad\quad\quad"Evidence":  "Ken, who is characterized as a murder suspect and a gambler.",\\
\quad\quad\quad"Relevant Paragraphs": 0\\
\quad\quad\}, \\
\quad\quad\{\\
\quad\quad\quad"Reasoning": "From paragraph 1, I know a character Ken was killed.",\\
\quad\quad\quad"Evidence":  "Ken is killed by Dixon during the confrontation, fitting the trope where a character with negative traits ends up being a victim.",\\
\quad\quad\quad"Relevant Paragraphs": 1\\
\quad\quad\} \\
\quad],\\
\quad"Answer":"yes" \\
    \} \\
Article: \{Given Article\}.Is the trope \{Given Trope\} related to the article?\\
   \bottomrule [1pt]
    \end{tabularx}
\end{table}

\subsection{Different CoT Analysis}

We intentionally designed two distinct binary CoT examples (refer to Appendix \ref{apdx:prompt}) to explore their influence on task outcomes. Despite this differentiation, our analysis (illustrated in Table \ref{table:diff_prompt}) revealed minimal observable difference in their outputs.






















\begin{table}[h]
\centering
\begin{tabular}{lccc}
\toprule[1pt]
 \textbf{Prompt} &  \textbf{F1} & \textbf{P} & \textbf{R} \\
\hline
 Example1 & 18.87 & 14.73 & 41.02\\
 Example2 &  18.03&  12.93 &   44.91 \\  
\bottomrule [1pt]

\end{tabular}
\caption{
ChatGPT results using two types of example prompts from Appendix \ref{apdx:prompt} with 10 random articles and 95 tropes.
}
\label{table:diff_prompt}
\end{table}



\section{Model Output Example}\label{apdx:output_example}

We automated the extraction of outputs from the LLM, where inputs will provide a JSON format example and the model typically follows this format for its responses (Figure \ref{CotExample}).

\section{LLaMa-2 Fine-Tune Detail}\label{apdx:finetune_detailed}

To address limited GPU memory, we employed a Parameter-Efficient Fine-Tuning (PEFT) technique called Low-Rank Adaptation (LoRA). This method focuses on adapting the most relevant model weights for the task, enabling efficient fine-tuning with fewer resources. Due to time and resource  
\\\\\\\\\\\\\\\\\\\\\\\\\\\\\\\\\\\\\\\\\\\\
constraints, we utilized the TiMoS training set, focusing on 20 tropes and a quarter of the total data for binary classification tasks. The multi-label task required 5 days of training, while the binary classification task took 30 days.

Given our resource limitations, the binary classification task involved training the model for 10 epochs, with movie synopses as input and "yes" or "no" as output. The multi-label task, also using movie synopses as input, aimed to identify all relevant tropes and required 20 epochs of training. We chose these epoch values based on the LLaMa2 paper \cite{touvron2023llama}, where their pre-trained model only needed 2 epochs for fine-tuning.

To assess performance, we selected the model with the highest F1 score on the validation set as the final result for testing. Due to the slow runtime of LLaMa2, we randomly sampled 100 synopses for validation from the original TiMoS validation set of 825 synopses.

\section{Attack Sentence}\label{aapdx:attack}

Table \ref{tab:attack_sentence} shows the example of the attack sentences used for the tropes in Table \ref{tab:attack} (Section \ref{subsec:4:10:attack}).

\lstdefinestyle{mystyle}{
  basicstyle=\ttfamily\footnotesize,
  breakatwhitespace=false,         
  breaklines=true, 
}


\lstset{style=mystyle}

\begin{figure}[h!]
\begin{minipage}{\textwidth}
\begin{parcolumns}{2}
\colchunk{\begin{lstlisting}[frame=tlrb]{Name}
{
    "Trope": "Adorkable",
    "Definition": "A character who is socially awkward or clumsy, but also endearing and lovable.",
    "Thought": {
        "Reasoning": "There is a character, Gary, who is portrayed as socially awkward and clumsy, but is also shown to be endearing and lovable.",
        "Evidence": "In paragraph 10, Gary is killed by the anaconda, leaving Denise devastated. This suggests that his character was portrayed in a way that made the audience care about him.",
        "Relevant Paragraphs": "10"
    },
    "Answer": "yes"
}
\end{lstlisting}}

\colchunk{\begin{lstlisting}[,frame=tlrb]{Name}
{
    "Trope": "Only Sane Man",
    "Definition": "A character who is the voice of reason and rationality in an otherwise chaotic or irrational situation.",
    "Thought": {
        "Reasoning": "There is no evidence or indication in the article that suggests the presence of a character who serves as the voice of reason or rationality in the chaotic or irrational situation described.",
        "Evidence": "N/A",
        "Relevant Paragraphs": "N/A"
    },
    "Answer": "no"
}
\end{lstlisting}}

\colplacechunks
\end{parcolumns}
\caption{Examples of ChatGPT's output(in JSON format), with responses of yes(left) and no (right)}
\label{CotExample}
\end{minipage}
\end{figure}

\begin{table}[h!]
    \centering
    \begin{tabularx}{\textwidth}{|c|X| } 
    \hline
     Trope & Attack Sentence \\ 
     \hline
     Kick The Dog & He and Jason are about to leave when they witness Kate and Glen arguing; when Glen shoves Kate to the ground, Rusty punches him in the face and flees with Kate and Jason.\\
     \hline
     Blatant Lies & Not wanting Mitch to realize the lengths to which she went to get his attention, she lies and tells him that her primary reason for coming to Bodega Bay was to visit Annie, a friend of hers from school. \\ 
     \hline
     Big Bad & During the tour around the facility, Walter sneaks into Kermit the Frog's office and discovers Statler and Waldorf selling the theatre to Tex Richman, an oil magnate, and his associates Bobo the Bear and Uncle Deadly. Once Statler and Waldorf leave, Walter learns of Tex's true intentions: to tear down the Muppet Studios and drill underneath for oil. Walter explains the situation to Gary and Mary, and the three track down Kermit at his mansion. \\ 
     \hline
    \end{tabularx}
    \caption{Example of the attack sentences}
    \label{tab:attack_sentence}
\end{table}



\end{document}